%% file: main.tex
\documentclass[letterpaper]{article}
\usepackage{aaai20}
\usepackage{times}
\usepackage{helvet}
\usepackage{courier}
\usepackage[hyphens]{url}
\usepackage{graphicx}
\usepackage[switch]{lineno}
\usepackage{subfig}
\usepackage{hyperref}
\usepackage{textcomp}
\usepackage{graphicx}
\usepackage{todonotes}

\newcommand{\gameR}{$\mathcal{R}$}
\newcommand{\etal}{et. al.~}

\begin{document}
  \title{Rinascimento: searching the behaviour space of Splendor}
   \author{Ivan Bravi, Simon Lucas\\
     \textit{School of Electronic Engineering and Computer Science}\\
     \textit{Queen Mary University of London}\\
     London, United Kingdom \\
     \{i.bravi, simon.lucas\}@qmul.ac.uk
   }
  
  \maketitle

  \begin{abstract} 
    The use of Artificial Intelligence (AI) for play-testing is still on the sidelines of main applications of AI in games compared to performance-oriented game-playing. One of the main purposes of play-testing a game is gathering data on the gameplay, highlighting good and bad features of the design of the game, providing useful insight to the game designers for improving the design. Using AI agents has the potential of speeding the process dramatically.

    The purpose of this research is to map the behavioural space (BSpace) of a game by using a general method. Using the MAP-Elites algorithm we search the hyperparameter space Rinascimento AI agents and map it to the BSpace defined by several behavioural metrics.

    This methodology was able to highlight both exemplary and degenerated behaviours in the original game design of Splendor and two variations. In particular, the use of event-value functions has generally shown a remarkable improvement in the coverage of the BSpace compared to agents based on classic score-based reward signals.
  \end{abstract}
  
  \section{Introduction}
    Most famous and modern applications of Artificial Intelligence in Games all seem to strive toward the same objective: show how humans are becoming obsolete. Often times however the conditions the two sides are subjected to are very different and fairness arises as a big question mark in the experiments. What is even more puzzling is the nature of the task through which the two are compared. Human beings can generally show an extremely wide selection of behaviours, even those that sometimes can be perceived as counter-productive or \textit{unintuitive} and these are exactly what we sometimes would classify as \textit{genius}. How can these traits be translated in the AI world and its lingo to beyond greedy AI players?

    In the world of academia, the effort in pushing AI applications in all direction is more uniform. Algorithms have been applied to content creation, game analytics and, of course, game playing. This last application has a number of vastly different declinations: we can have the renowned \textit{performance game-play}, where the objective is beat some state of the art algorithm or human grand master; another example is \textit{believable game-play}, where the AI is designed to resemble the skills and behaviours of a fellow human being; and also we have AI for \textit{play-testing}, where agents are used as a cheaper and less biased alternative to human beings for testing and early-iteration-stage version of a new game.

    Play-testing AI can be approached extremely precisely in certain scenarios, testing the solvability of a level in puzzle game like Sokoban is clearly measurable and easily implemented using well known algorithms like A*. On other occasions play-testing involves more nuanced evaluations of human-centred game-play aspects that don't necessarily map to generic measurable quantities. One example is \textit{behaviour} which is characteristic of each game and needs clearly some domain-specific knowledge to be defined.

    In the last decade several research projects have focused on the development of the so called Artificial General Intelligence (AGI), which has the purpose of using as little domain knowledge as possible (if not none) in the pursuit of its goal. This was particularly noticeable in Game AI where AGI-oriented frameworks were developed to support this effort. Techniques have been developed in both learning and planning scenarios, in the first the AI has the opportunity to gradually learn how to play the game during a training period, in the latter a forward-model is provided through which the AI can simulate actions leading to future states.

    In this generic setting, developing play-testing techniques is extremely challenging especially if the objective goes beyond evaluating simple concepts like solvability. This is where this research takes place, with the use of AGI planning techniques and newly developed Event-value Functions (EF), we show how to enhance (w.r.t. other classic approaches) the ability the AI, that we will call EF-AGI, to embodying varied and multi-faceted behavioural traits, and explore and characterise the performance profile of a behavioural space relative to the game. This exploration is carried out using the Multi-dimensional  Archive of Phenotypic Elites (MAP-Elites) algorithm, a quality diversity algorithm for Evolutionary Computation. Behaviours will be captured by metrics crafted by a designer which is a much simpler task than guessing a-priori the many possible ways a game could be played.
    Summing up, the search for possible behaviours will be carried in the hyperparameter space of an EF-AGI agent and mapping its instances to the behavioural space defined by the behavioural metrics.



  \section{Background}\label{sec:background}

    The research surrounding this work comes from several different areas: AI behaviour characterisation, AI play-testing strategies, Evolutionary Algorithms and Statistical Forward Planning (SFP).

    \subsection{Rinascimento}
      The Rinascimento framework (\gameR) \cite{bravi2019rinascimento} \footnotemark~was developed around the board game Splendor\texttrademark designed by Marc Andr\'{e} and published by Space Cowboys. The game takes place in the Renaissance and the players' objective is gathering wealth and the respect of the nobles. In order to achieve this, they can pick tokens from the table and use them to buy cards which are worth points and can attract the nobles, also worth points. Once a player has reached 15 points the game is over.
      \gameR~implements a flexible game engine that allows the user to alter the rules of the original game in a parametric way, e.g. increasing/decreasing the maximum number of tokens a player can have. Further more it allows to add and remove mechanics, use custom decks of cards and nobles. This required a flexible way of dealing with the action space which can be sampled randomly by querying the game state. This feature is crucial also for dealing with the combinatorial nature of the game (see \cite{bravi2019rinascimento} for more details) where listing all the legal actions would have been computationally expensive.
      Finally the AI agents that come with the framework are highly customisable in their hyperparameter and the heuristics they use to guide the search in the action space.
      The algorithms implemented, Rolling Horizon Evolutionary Algorithm (RHEA) and Monte Carlo Tree Search (MCTS), fall in the category of SFP methods, these are planning algorithms that, using a forward-model of the game, simulate actions and future game-states so, by using a heuristic that evaluates the quality of a state, they can select their next action to play. Such heuristic is a crucial component of the agent since it has the power of affecting the agent's in-game behaviour.

      \gameR~implements the recently introduced Event-value Functions \cite{bravi2020rinascimento} with the purpose of providing a much more precise characterisation of AI's behaviour compared to a classic score-based function, details in Section \ref{sec:event}. This required a logging system attached to the game engine and keeps track of all the events taking place in the game.

      In conclusion the \gameR~was developed as a workbench for \textit{exploring} the neighbour of Splendor-like games, hoping to promote procedural content generation and a thorough approach to hyperparameter tuning of the AI players.

      \footnotetext{github.com/ivanbravi/RinascimentoFramework}

    \subsection{Characterising AI game-playing style}
      One extremely relevant aspect of game-playing AI, but unfortunately rarely confronted directly, is that of controlling and characterising the agent's style. Is it aggressive or defensive? Does it gather plenty of resource? Is it curious and explores everything? These general questions can get much more precise and detailed once the game genre is defined and even more when a single game is picked.
      But what is the effort put by the designers of AI algorithms into approaching this topic?

      From the perspective of solely measuring these behavioural characteristics an effort was made to define a set of generic metrics \cite{volz2017gameplay} that can be potentially applied to any game. These can be as simple as duration and score of a game to more complex aspects of game-play such as action entropy and agent decisiveness (also explored in \cite{bravi2018shallow}). These measures can also be of abstract concepts as drama and surprise. However, this approach is still in its early days and a more domain-specific approach is common, custom metrics are usually developed by the researchers or game-designers in order to highlight the desired features.
      
      When it comes to enabling an AI algorithm to embody different strategies H\"olmgard \etal have explored multiple approaches to identify the so-called procedural personas, AI agents modelled around Bartle's player types. In \cite{holmgard2018automated} the procedural personas were based on a MCTS agent with customised tree policies equations evolved using Genetic Programming. The four personas (runner, monster killer, treasure collector and completionist) are based on custom designed utility functions used in the rollout phase of MCTS. The tree policies evolved combine together game-play metrics custom designed for the rouge-like game \textit{MiniDungeons 2} such as: potions drunk, javelins thrown, traps spring etc. This idea of measuring the interaction with the elements of the environment is grounded in psychological decision theory, by the definition of the concept of affordance: the ability of interacting and manipulating an object. The four personas were tested on several levels, the results highlighted the differences between their play-styles but also how each one is suited for different levels. The only persona that didn't show its superiority in any level was the completionist which appeared as an inferior version of the treasure collector. However, the completionist was still able demonstrate its characteristic behaviour when compared to the Runner by consistently completing all the levels and interacting with more game elements.

      Finally, in \cite{bravi2020rinascimento}, the authors introduce the concept of Event-value Functions. Inspired by State-value Functions (SF), which are wide spread in the Machine Learning applications for Game AI, and \textit{affordance} these functions work as heuristics for the game dynamics based on event logging. By monitoring the events triggered during an action sequence, EFs synthesise and event-log into a feature vector of fixed size, this is then fed to a mixing function that can be tuned by adjusting its weights.
      In this context an \textit{event} is any modification to the game state whether this comes from a player interacting with the environment, or within the environment itself. Events are characterised by a set of attributes, the full list is available in \cite{bravi2020rinascimento}, for this work were used: \textit{type id}, unique incremental integer identifier of the event type; \textit{who}, specifies the subject that triggered the event. These attributes are used to filter the events triggered by the player using the EF and by counting the event type ids to produce the feature vector. The authors explored several mixing functions of varying complexity, namely, linear and polynomial models, they also experimented with two ways of mapping type ids: the default one by using all 18 events and a hand-crafted one that discards and groups the event in 5 macro-events.
      The main purpose of their research was to create a heuristic framework, that allowed AGI algorithms to characterise more precisely their game-play behaviour. The results show how this objective was achieved by also improving their performance compared to the current state of the art score-based heuristic AGI agents. This was possible even if EF didn't use any measure of score variation, but just a flag signal when the score was increasing/decreasing.

    \subsection{AI for playtesting}
      The use of AI agents for playtesting has been explored in both academia and the industry.

      In the industry we have seen several approaches used for playtesting games and highlighting potential flaws of the game design. In \cite{silva2018exploring} the authors show how a cut-down version of the game The Sims Mobile was used to speed up the simulation process in order to provide feedback on unbalanced elements in the gameplay, inconsequential rewards and a general evaluation of strategic effectiveness. The well-known $A^*$ algorithm was used to play-test the game and gather information about the game play in the form of game-play metrics. The results were used by the designers to change the game rules and improve the players experience.
      Another set of games where AI playtesting was adopted is Candy Crush Saga and Candy Crush Soda Saga \cite{gudmundsson2018human}, the authors trained a Convolutional Neural Network (CNN) to predict the most human-like actions based on human data and used the CNN to play-test levels and predict their difficulty before being released to the players in the real game.

      In academia we can find domain specific applications as \cite{de2017ai} where the authors used custom designed AI agents to analyse the popular board game Ticket To Ride.

      Other research projects instead, have focused on producing design tools with the objective of using AI to quickly test levels and provide feedback to the designer. In an AGI-oriented approach, Machado \etal\cite{machado2018ai} have created an AI assisted tool for designing and debugging 2D sprite-based arcade-style games. The tool is based on the General Video Game AI (GVGAI) framework and uses the Video Game Description Language (VGDL) for fast prototyping games. It provides several useful features for effective game designing but most importantly a set of AGI players (mostly implementing SFP techniques) for ready to use agents.
      
      Promising work was done by Zook \etal\cite{zook2019automatic} where a method based on Active Learning was used to speed up the play-testing phase of a shoot-‘em-up game by defining and modelling its hyperparameter space. The idea is that as new versions of the game are played, the AI learns how the hyperparameter space maps to a fitness value representing the quality of the design. This model is then used to suggest new versions of the game to play-test. This process, however, still requires human players and is more appropriate for a final tuning of the game.

      In the work of Guerrero \etal\cite{guerrero2018using} the idea of using a team of AGI agents to support game design is explored by suggesting a team of agents with set behavioural characteristics. These are described in a generic fashion supposing general features of the game developed such as a score, Non-Player Characters to kill and items to collect.

      Most of these works either rely on given AGI agents or agents custom-designed for a game, thus obtaining feedback that is either generic but potentially shallow or specific but requiring effort in creating tailored agents for a single game.

      In this piece of research instead we will flip the problem by asking the question: what are the different behaviours that our agents can embody? This has the all the virtues of a general approach without the limiting factor of relying on pre-defined behaviours. This will be possible thanks to the considerably vast hyper-parameter spaces defined by EF-based AGI agents. It also give back to the designers the control on which behaviours to focus their attention.

    \subsection{Quality Diversity: MAP-Elites}

      In evolutionary computation, the concept of Quality Diversity (QD) \cite{pugh2016quality} has arisen in recent years to resemble more closely the natural processes of evolution where the "search" for fit individuals in an environment is carried on in parallel across multiple species that develop extremely divergent phenotypes. QD is a category of algorithms that focus on maintaining a \textit{diverse} selection of individuals striving together for improving their \textit{quality}.

      Since its inception the  MAP-Elites \cite{mouret2015illuminating} algorithm has appeared as a remarkable way not just to search spaces with the purpose of optimisation but also for \textit{illuminating} the space. 
      Given a high-dimensional search space, the algorithm maps it to a low-dimensional feature space (diversity) plus a performance measure (fitness in EC lingo) representing the best point found so far (quality) . This serves the purpose of lowering the complexity of the search space into a more high-level and interpretable space where each feature coordinate can be characterised by its fittest individual.
      Generally speaking the concepts of feature or behaviour (and similar, e.g. feature space or behaviour space) are considered equivalent in this setting. In the following we will prefer the term \textit{behaviour} as better fitting the context.

      Given a search space, a behaviour space discretised in buckets by the user, an evaluation function producing a fitness value and a behaviours vector, the algorithm is divided in two loops: the booting loop and the main search loop. During the first, $N^{boot}$ individuals are sampled and evaluated from the search space. Instead, during the search loop, individuals from the BSpace are randomly sampled, mutated and evaluated until $N^{budget}$ evaluations are performed.
      After the evaluation the individual is assigned to its bucket in the BSpace according to the behaviours vector, if the bucket is already filled the individual with the highest fitness is kept.

      In Game AI we have seen already several applications of the algorithm, most of them seem to focus on procedural content generation.
      In \cite{khalifa2018talakat}, the authors search the space of levels for a bullet hell game. The space is defined through a context-free grammar describing the features of the elements (e.g. bullet spawners and bosses) in the level, moreover, each description was constrained through gameplay metrics. The MAP-Elites algorithm was enhanced with a feasible/infeasible 2-population mechanism that preserved also illegal (according to the above constraints) descriptions in the hope of saving valuable genetic material that mutation can eventually fix. Levels were evaluated by A* agent whose actions were restricted in order to influence its dexterity and strategy, 9 different combinations were tested returning a variety of levels suitable to each agent.

      Another application but on a completely different game is \cite{fontaine2019mapping}, where space of decks in Hearthstone was searched. In this work MAP-Elites is augmented with Sliding Boundaries (MESB), meaning that the buckets boundaries of the BSpace can be adjusted to reflect the population density. This favours a more fair local competition between individuals and a more homogeneous illumination of the BSpace. The results have shown that MESB was able to find high-performing decks in different strategy spaces.

      More recently \cite{gonzalez2020finding}, the authors employ the MAP-Elites algorithm within an Intelligent Trial-and-Error procedure to find levels with the desired level of difficulty for a target AI agent. The levels are play-tested using a different set of algorithms whose performance results is fed to a Bayesian Optimisation algorithm to predict the performance of the target agent. This approach has shown promising results by reliably providing good results within very few iterations.

      To the best of the authors' knowledge, this is the first piece of research focusing on exploring and illuminating the behavioural space of AI agents.

  \section{Games}
      In this research we use three games based on Splendor leveraging the \gameR~framework.
      Splendor is a competitive multiplayer card-based board game, on the table there are 3 decks of cards with 4 cards face-up for each deck, 5 stacks of common tokens of different colours, 1 stack of joker tokens and noble tiles. Players play performing a single action per turn, they can either: \textit{pick three tokens} of three different suits;\textit{pick two tokens} of the same suit, but only if there are more than 3 tokens of that same suit; \textit{reserve a face-up card} from the table and get a joker token if available; \textit{reserve the top card from one of the decks}, the other players can't see the card until the player buys it; \textit{buy a card from the table}; \textit{buy a reserved card}.

      Players can't have more than 10 tokens in their hand and they can't have more than 3 reserved cards at a time. In addition when they buy a card the cost is discounted by the amount of cards in the player's hand of each suit.
      Nobles move to the player's hand once per turn whenever the player has enough cards as specified by the noble tile. Once a player reaches 15 points the round finishes and the winner is the player with more points or, in case of draw, less cards
      The three games are:
      \begin{itemize}
        \item 2-Player Splendor (SP2P): 5 tokens per suit and 3 nobles.
        \item Wacky Splendor (W2): several game parameters have been tweaked in order to potentially allow wildly different player behaviours. There are 10 tokens for each suit, the token limit for the players is 20, 1 single noble. The action set is also slightly different, the player can pick two coins of two different suits or three of the same one and without restrictions.
        \item 1 card to win (1C2W): the decks have been redesigned to introduce few very expensive cards worth 15 points so that the first player to buy one wins the game. All the other cards have very few points so that it's not possible to win without a 15-points card.
      \end{itemize}

      In addition to the previous work on \gameR~we introduce and encoding of game state so that we can compare EFs and SFs.
      The state encoding is built hierarchically packing together information following this grammar: State$\rightarrow$Board [Player]; Board $\rightarrow$ [Deck][Noble] TokensAmount; Deck $\rightarrow$ CardsRemaining [Card]; Card $\rightarrow$ Suit Cost Points; Noble $\rightarrow$ Points Cost; Player $\rightarrow$ Points TokensAmount CardSuitCount [Card]. \textit{CardsRemaining} is the counter of cards left in the deck. \textit{TokensAmount} is a vector of the amount for each token type (order is consistent). \textit{Suit} is the one-hot encoding with as many bits as suits in the game. \textit{Cost}, similarly to \textit{TokensAmount}, is a vector of the cost for each token type. \textit{CardSuitCount} is the amount of cards, divided by suit, that a player has bought. This give us a representation of a SP2P game state in 144 dimensions which dwarfs the 5- and 18-dimensional representation used by EFs.

    \begin{figure*}[t]
      \centering
      \includegraphics[width=\textwidth]{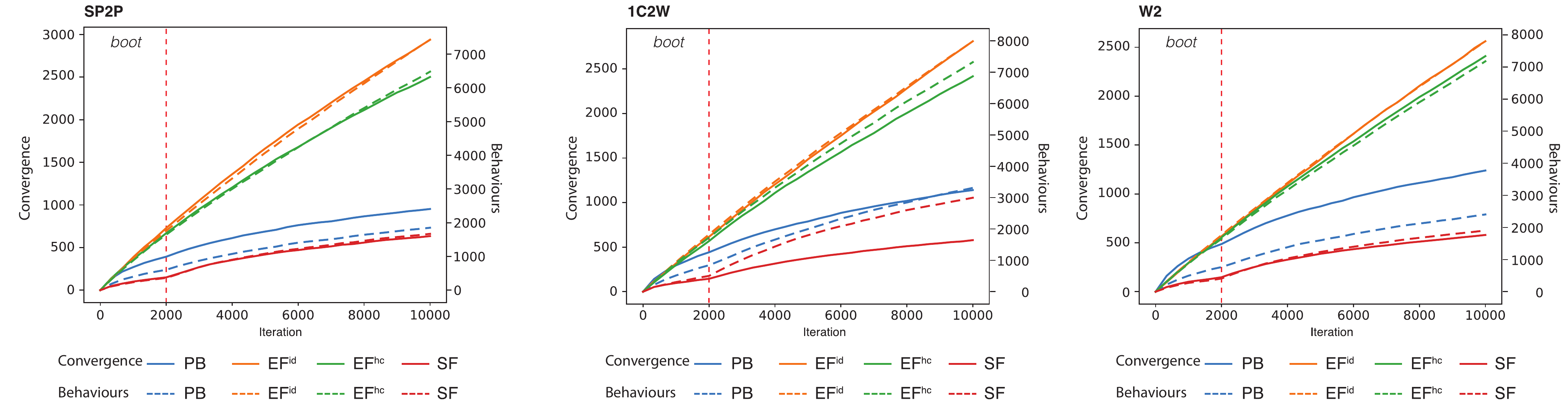}
      \caption{\label{fig:conv_exp}Picture showing the convergence (solid lines) and the exploration (dashed lines) for the three different games.}
    \end{figure*}

    \begin{figure}[t]
      \centering
      \includegraphics[width=\columnwidth]{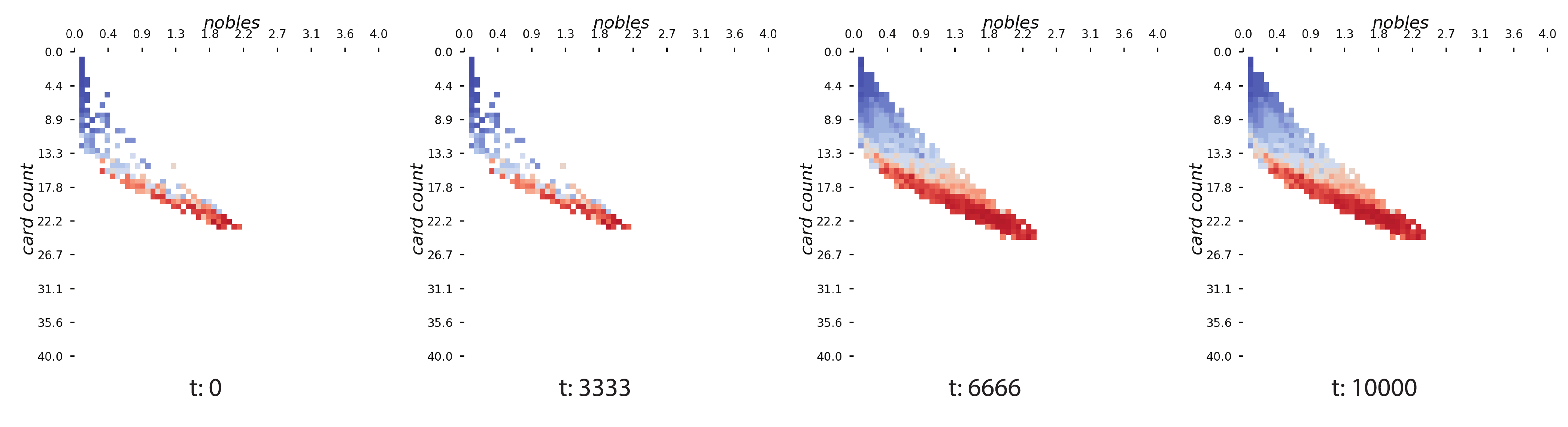}
      \caption{\label{fig:gif}One of the heatmaps of the EF$^{id}$ search in SP2P as interation count t advances.}
    \end{figure}

  \section{Agent Hyperparameters: Search Space} \label{sec:event}

    The search space used by MAP-Elites is defined by the hyperparameters of the Branching Mutation Rolling Horizon (BMRH) agent which was used also in \cite{bravi2020rinascimento}. This agent, based on the Rolling Horizon Evolutionary Algorithm \cite{gaina2020rolling}, evolves a sequence of \gameR~actions and implements a special mutation operator that instead of mutating a single action in the sequence, given a random starting point, it rolls a whole portion of the sequence so that all the random actions substituting the previous ones are legal.

    In this research we use the BMRH agent in combination with several heuristic:
    \begin{itemize}
      \item Point Based (PB) heuristic: returning the score delta between the beginning and the end of the sequence.
      \item EF heuristic: analyses the events logged during the sequence.
      \item SF heuristic: evaluates the quality delta of the state reached compared the starting one.
    \end{itemize}

    As seen previously, with an EF we can use a number of different mixing functions (or models): linear, polynomial, neural networks and so on. For the sake of simplicity we decided to use only a linear function as this has shown comparable peak performance to PB-BMRH. Similarly the linear function was adopted also for SFs, in fact the state-value functions feed the state encoding to a model in order to compute a scalar value representing the quality of a game state.

    The hyperparameter spaces searched are a combination of the base BMRH hyperparameter space (see \cite{bravi2019rinascimento} for a full description) characterised by 10 dimensions and the hyperparameter space of the heuristic function. Since we are using a linear function, the size of each space will be equal to the size of the features used by each heuristic.
    The individual size for the four spaces are: PB, 10D; EF$^{id}$, 28D; EF$^{hc}$, 15D; SF, 154D.

  \section{Metrics: The Behavioural Space} \label{sec:metrics}
    Behaviours are an high-level concept that we can formalise by measuring it through clearly defined and objective metrics. Since the evaluation of a behaviour needs to be gathered through multiple trials, the final value vector $\mathbf{b}$ of metrics is obtained by averaging each metric across the trials. The BSpace is then created by first picking the boundaries of each metric and then by uniformly dividing the range into a predetermined amount of buckets. In particular, the first and the last bucket in each dimension contain also all the behaviours exceeding the boundary limit.
    For this work, we defined 5 different metrics that can help describing different manners of playing Splendor, these are intended as player-centric, relative to a single player and recorded during a single game (boundaries and bucket count are reported between parenthesis):

    \begin{itemize}
      \item \textit{card count}: how many cards were bought ($[0,40]$; $62$);
      \item \textit{total coins}: the amount of tokens picked ($[0,120]$; $62$);
      \item \textit{nobles}: how many nobles the player attracted ($[0,4]$; $62$);
      \item \textit{card cost}: the mean cost of the cards bought ($[0,15]$; $62$);
      \item \textit{reserved cards}: the amount of cards reserved ($[0,30]$; $62$);
    \end{itemize}

    This set of metrics, was picked as representative by the authors, however, depending on the purpose of the play-testing process, a designer might decide to focus on a particular aspect of the game with more nuanced and specific metrics.

    The performance metric used as fitness by the MAP-Elites algorithm is the win rate computed over several playthroughs.

    \begin{figure*}[t]
      \centering
      \includegraphics[width=\textwidth]{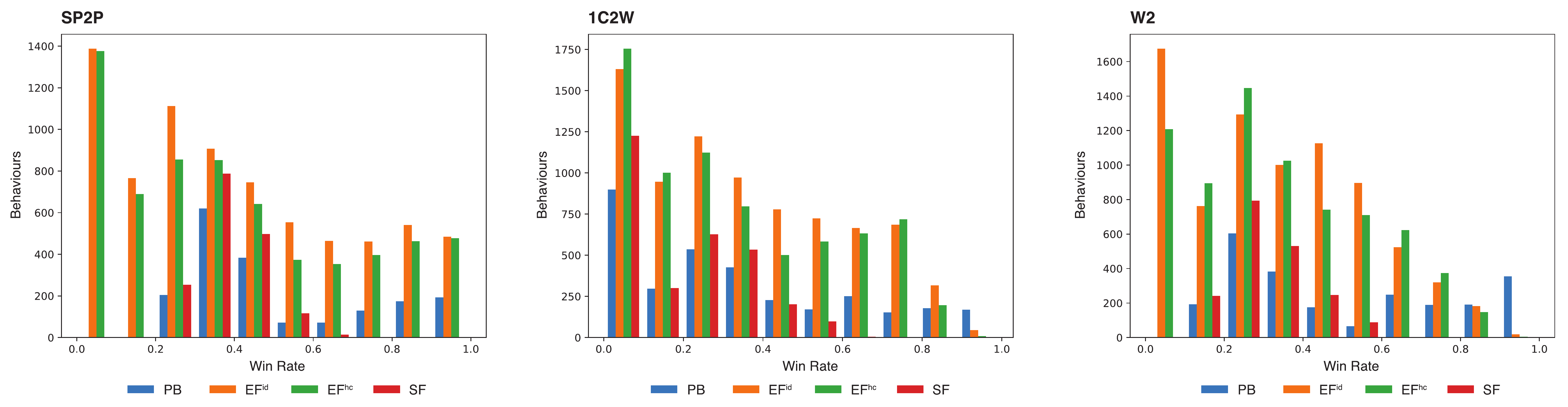}
      \caption{\label{fig:performance}Picture showing the performance distribution for the three different games.}
    \end{figure*}

  \section{Experiments} \label{sec:experiments}

    The purpose of these experiments is to two-fold. (1) Provide an understanding of what is the most appropriate type of heuristic between PB, EF and SF to be used when exploring the behavioural space of a game. (2) Provide a set of intuitive visualisations that can help the game designer understanding the features of the BSpace of a specific game.

    \subsection{Experimental Details}
  
      The Kotlin implementation\footnotemark~of the MAP-Elites algorithm was run for a total of 10000 evaluations and with a booting process of 2000 samples. For each experiment we fixed: the game (SP2P, W2, 1C2W), the search space (PB, EF$^{id}$, EF$^{hc}, SF$) and the opponent (random, BMRH*); for a total of 12 experiments. Each evaluation consisted of 100 games played independently and in parallel over 8 threads on the \textit{omitted} High Performance Computing services \cite{king_thomas_2017_438045_omitted}. \footnotetext{github.com/ivanbravi/MAP-Elites}
      The number of samples was picked to allow for a fast execution without compromising too much the reliability of the results. This amount of samples allows us to have a margin of error with 95\% confidence of $\pm 10$ win percentage, given the formula $1.96 \sqrt{p(1-p)/100}$ in the worst case scenario of $p=0.5$. The size of this range is by no means a limitation since the purpose of the experiments is evaluating the global trends in the BSpace.
      The computational resources (10 cores) allocated were limited in order to consider a scenario where the designer has available just an average computer.

      The values composing the hyperparameter space of the heuristic's weights consisted in 11 samples taken uniformly in [-1, 1]: \{-1, -0.8, -0.6, -0.4, -0.2, 0, 0.2, 0.4, 0.6, 0.8, 1\}.

    \subsection{Analysis Methodology}

      Since the BSpace is defined by 5 behavioural metrics and a performance metric, it is crucial to analyse the data both through simplified representations. This is particularly important when we try con compare the results obtained under different experimental conditions.

      During each execution we tracked the BSpace population iteration by iteration, this allows us to analyse not only the final results of the experiment but its whole evolution.

      The main consideration when it comes to visualize 6D data in a comprehensible way. Most example mentioned in Section \ref{sec:background} visualise a 2D heatmap by selecting two behavioural dimensions and averaging across the others and using the performance metric as heat value.
      By fixing two behaviours, averaging is just one of the many possible approaches in analysing the remaining dimensions, e.g.: variability, difference between minimum and maximum; risk, likelihood of good performance; density, purely amount of different behaviours.
      In this work we filter the \textit{max} value since we are interested in what is the most favourable outcome for following a certain behaviour. This also means that all the peaks in different plots are relatable.

      \subsubsection{Single Space Analysis}
        Visualising the BSpace is the first step, thus for each unordered couple of behaviours $X$ and $Y$ we project BSpace's remaining dimensions on the plane $XY$ using win rate as heat. This gives us 10 heatmaps contributing to purpose (2).
        Another way of looking at the data is to keep track of how many different behaviours are in the BSpace at each iteration, since one of our objectives is illuminating the behavioural space. This metric will explicitly measure the degree of coverage of the BSpace thus a comparison measure of success/failure for (1).
        Measuring the quality of the data gathered is not an easy task. The results could have high coverage of the BSpace but with poorly performing behaviours or low coverage overall high performing. We decided to use two separate metrics. The first is a performance histogram, it roughly shows how many behaviours we have found along the performance metric axis. Since the statistical reliability of the performance metric is not extremely precise, it's not worth binning the data too finely, we will use 10 bins. Seeing the distribution from bad to good behaviours is relevant to both (1) and (2) since it provides a general idea of how many behaviours per bin were found and where is shifted the distribution, e.g. too many good behaviours suggest a trivial game. The second is an analysis of the convergence of the algorithm, it consists in the sum of the performances of all the unique behaviours in the BSpace as budget is consumed. This quantity is related to convergence since it is a monotonic bounded function, with initial (lower-bound) value of 0 and maximum (upper bound) value equal to the sum of all the possible best performance per behaviour. This is useful to (2) when evaluating the amount of computation dedicated to the experiment.

      \subsubsection{Two Space Comparison Analysis}
        All the previous plots, with exception of the heatmaps, can be aggregated in a single plot allowing for straightforward comparison of the results.
        Since the BSpace is exactly the same for all the experiments we can overlap them and produce two different kind of plots. Let's suppose we are comparing experiments A and B we can produce a \textit{coverage} plot showing areas are covered by both A and B, A but not B and the converse, this will provide an immediate feedback on what experiment was able to illuminate better the BSpace. We can also go into more detail by plotting a \textit{performance delta} plot, by subtracting the results of experiment A from those of B. This will show the behaviours where one agent type is stronger than the other.
    \begin{figure*}[!t]
      \centering
      \includegraphics[width=\textwidth]{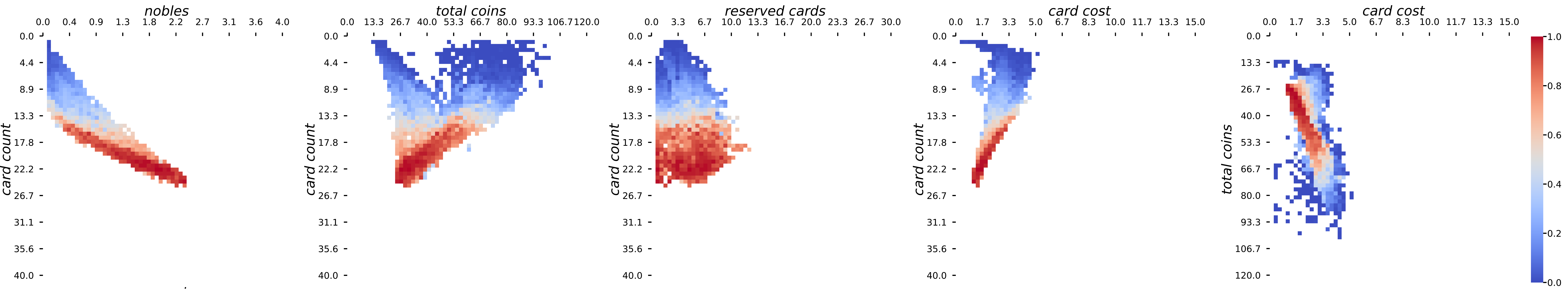}
      \caption{\label{fig:exploration}Picture showing a selection of the heatmaps for EF$^{id}$ exploring the SP2P game.}
    \end{figure*}

    \begin{figure}[!t]
      \centering
      \subfloat[coverage]{
        \label{fig:singles:cover}
        \centering
        \includegraphics[width=.48\columnwidth]{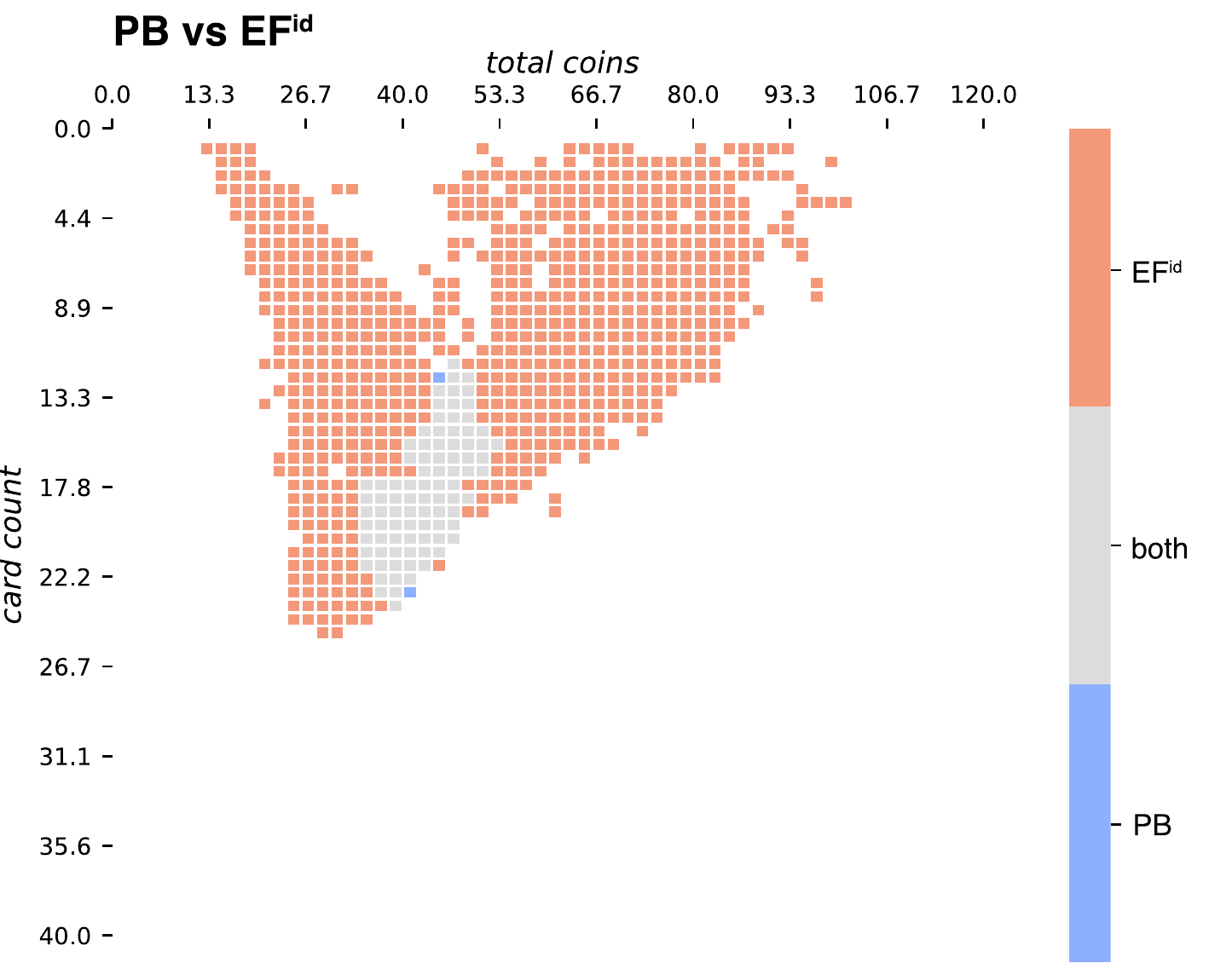}
      }
      \subfloat[performance delta]{
        \label{fig:singles:delta}
        \centering
        \includegraphics[width=.48\columnwidth]{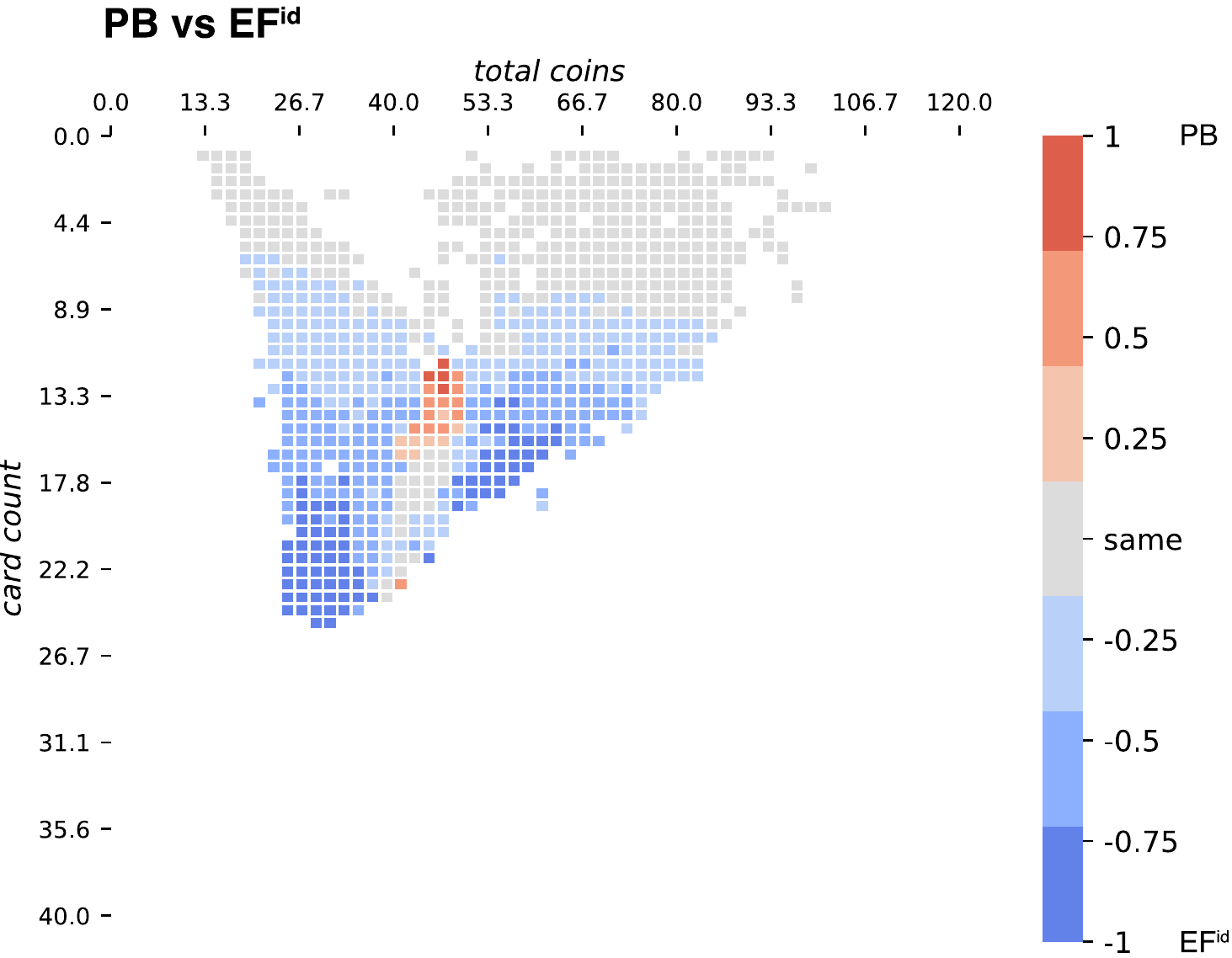}
      }
      \caption{\label{fig:singles} The heatmaps show \textit{coverage} (\ref{fig:singles:cover}) difference and \textit{performance delta}(\ref{fig:singles:delta}) between PB and EF$^{id}$ in SP2P.}
    \end{figure} 

    \begin{figure}[!t]
      \centering
      \subfloat[SP2P]{
        \label{fig:bug:sp2p}
        \centering
        \includegraphics[width=.48\columnwidth]{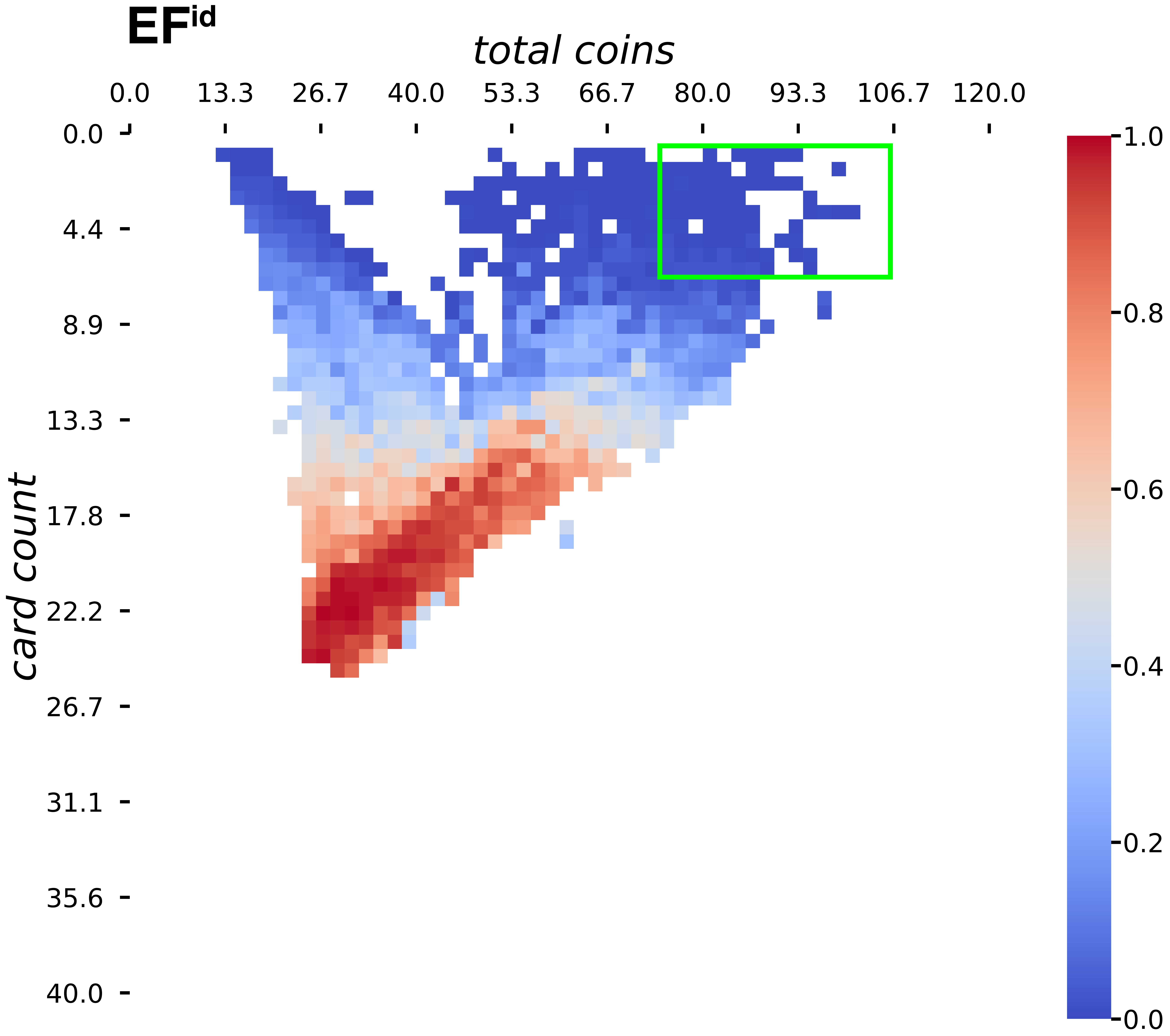}
      }
      \subfloat[W2]{
        \label{fig:bug:w2}
        \centering
        \includegraphics[width=.48\columnwidth]{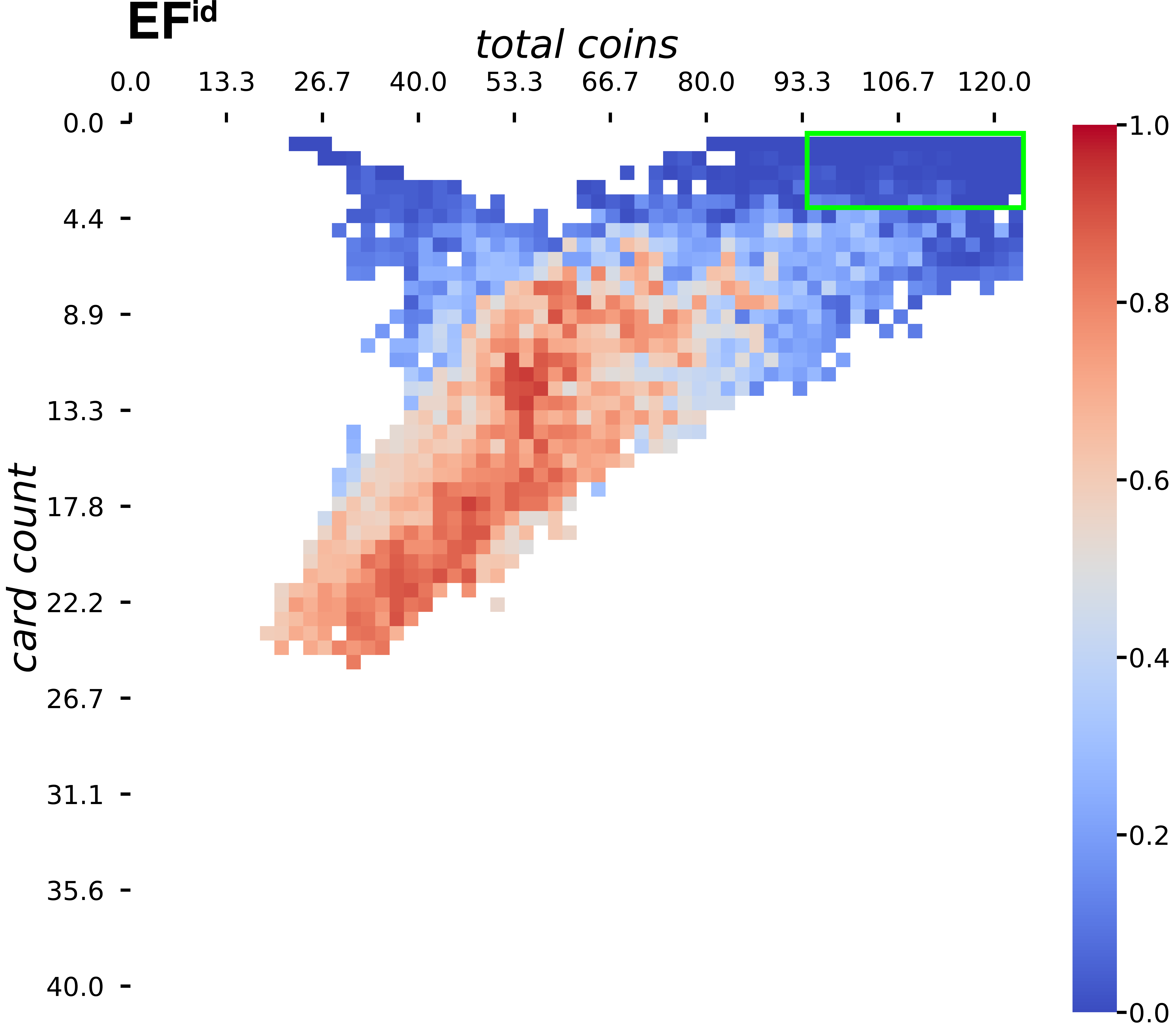}
      }
      \caption{\label{fig:bug} The heatmaps show in the green areas the same degenerate behaviour in SP2P (\ref{fig:bug:sp2p}) and W2 (\ref{fig:bug:w2}).}
    \end{figure} 

  \section{Discussion} \label{sec:discussion}
    Given the amount of experiments and the complexity of the data to analyse we will show only a limited amount of plots, the full collection of analysis plots is available online\footnotemark. \footnotetext{\href{http://www.github.com/ivanbravi/RinascimentoFramework/tree/master/Media/Papers\%20Plots/AAAI}{/RinascimentoFramework/Media/Papers Plots/AAAI}}
    A single experiment took between 10 and 74 hours, from running overnight to over the weekend and Monday, the duration was mainly a function of the game version.
    Looking at the convergence plots in Figure \ref{fig:conv_exp} for all three games we can notice that the potential for further exploration of the space is still very high, however when following the search progress on the heatmaps we could see that the 2D space coverage had almost converged, see an example in Figure \ref{fig:gif}. Another striking fact is how the four methods always rank in the same order for amount of behaviours found: EF$^{id}$, EF$^{hc}$, PB, SB. In particular the gap from the two EFs seems to be quite limited meaning that the hand-crafted grouping was likely adequate. What is surprising, instead, is how SF yields such a poor result. This is likely due to the extremely bigger search space and the amount of one-hot-encoded data that in a linear model is probably of little help. We can also see how SF and PB are almost identical in the amount of behaviours illuminated, what separates them is the performance of those points.
    Figure \ref{fig:exploration} shows the non-trivial behavioural landscapes illuminated by EF$^{id}$. We can notice for example the importance of getting both cards and nobles (first heatmap on the left) and how after around 15 cards, nobles start to naturally increase and with that the win rate. Another feature of the game represented (third heatmap from the left) is the low impact of reserved cards, which in Splendor is important mainly to block the opponent.
    By looking at the performance distributions in Figure \ref{fig:performance}, we can see how generally the two EF heuristics are able to cover much more behavioural space than the both PB and SF. This seems to be also true in the specific case of well performing behaviours, with the only exception of the game W2. This is likely due to a game version where, given the higher availability of coins, is more susceptible to being able to perceive score variation to which EF are partially blind to, whereas PB excels at.
    
    Figure \ref{fig:singles} shows how two experiments can be overlapped, we see here how EF$^{id}$ is clearly superior in coverage to PB (with only 2 exclusive behaviours) and, except for a small pocket, also in terms of performance. PB can sometimes be superior since it has a finer understanding of score variations.
    In Figure \ref{fig:bug} we can see how degenerate behaviours can arise from the excessive use of specific mechanics of the game: in this case we have agents that during the game reach their maximum amount of coins and keep swapping coins instead of spending them for buying cards.



  \section{Conclusions and Future work} \label{sec:conclusion}
    In this work we have presented a novel approach to playtesting games where the objective is to provide the game designer with a representation of the behavioural space of a game. The methodology uses the MAP-Elites algorithm to illuminate such space by searching the hyperparameter space of an AGI SFP algorithm working as configurable play-tester agent. Three different versions of the game Splendor were tested using \gameR~and comparing four different types of agents: score-based BMRH, Event-value Function BMRH (with hand-crafted and with identity id mappings) and State-value Function BMRH.

    EF-BMRH has shown remarkably superior abilities in illuminating the behavioural space for all the three games. Overall, EFs appear the most flexible heuristic model when it comes to characterising gameplay behaviour, which was the original objective. The search is able to highlight the most promising behaviours (tactics) for playing the game providing extremely valuable information to the designer. It also highlighted degenerate behaviours suggesting possible areas of design improvement. We can say to have succeeded in both objectives (1) and (2), although the full potential will be expressed once in the hands of game designers.

    An obvious application for future work would be to use more sophisticated function models beyond the simplistic linear. Artificial Neural Network sound like a perfect fit for the application, however keeping the number of hyperparameter as low as possible will be a crucial aspect for not losing feasibility in a real play-testing setting. This will hopefully be a more appropriate model for SF. The methodology could be ported to an AGI-oriented framework such as \mbox{GVGAI} to test a bigger set of more varied games.

  \section*{Acknowledgements}
  	\addcontentsline{toc}{section}{Acknowledgements}
    This work was funded by the EPSRC CDT in Intelligent Games and Game Intelligence (IGGI) EP/L015846/1.
  	This research utilised Queen Mary's Apocrita HPC facility \cite{king_thomas_2017_438045}, supported by QMUL Research-IT.

\bibliography{bib.bib}
\bibliographystyle{aaai}

\input{Appendix}

\end{document}

%% file: Appendix.tex
\setcounter{secnumdepth}{2}
\newcommand{\n}{n\textdegree~}
\newcommand{\param}[1]{\textit{#1}}

\newpage
  \appendix
  \section*{Appendix}
  
  In Section \ref{sec:rinascimento} we provide further information about the game parameters, the specifications of the two game versions introduced, the hyperparameters of the BMRH algorithm and, finally, the events used by the Event-value Function heuristics.

  In Section \ref{sec:exp} we describe the supplemental material provided: data resulting from the experiments, the scripts used to analyse it and the plots produces.
    
\section{Rinascimento}\label{sec:rinascimento}
    \subsection{Game Parameters}
      The idea of Rinascimento is to have a game engine that can easily run Splendor-like games. This is made possible by a fully parametric implementation. Each game specification can be loaded through a \textit{json} file specifying the parameters. In Table \ref{table:setup} are listed the parameters regarding the set up phase of the game. Instead, Table \ref{table:rules} lists the parameters related to the rules of the game. Finally, in Table \ref{table:actions} you can find the parameters defining how the actions are defined.
      \begin{table}[h]
          \centering
          \begin{tabular}{l|c|c}
          \textbf{Description}     & \textbf{Symbol}  & \textbf{Default}   \\
          \hline
          \n players              &\param{P}          & 4         \\
          token types*             &\param{nTT}        & 5         \\
          \n joker token          &\param{nJT}        & 5         \\
          \n decks*                &\param{D}          & 3         \\
          \n face-up cards        &\param{FUC}        & 4         \\
          \n extra noble*          &\param{EN}         & 1       
          \end{tabular}
          \caption{Parameters extracted from the game's setup. The ones marked with a star require PCG.}
          \label{table:setup}
      \end{table}

      \begin{table}[h]
          \centering
          \begin{tabular}{l|c|c}
          \hline
          \textbf{Description}        & \textbf{Symbol}   & \textbf{Default}   \\
          \hline
          max \n tokens per player    & \param{maxT}      & 10        \\
          max \n reserved cards       & \param{maxRC}     & 3         \\
          end-game prestige points    & \param{PP}        & 15        
          \end{tabular}
          \caption{Parameters extracted from the game's rules.}
          \label{table:rules}
      \end{table}

      \begin{table}[h]
          \centering
          \begin{tabular}{l|c|c}
          \hline
          \textbf{Description}                                    & \textbf{Symbol}   & \textbf{Default}   \\
          \hline
          \n different token types in \textit{pick different}     &\param{nTTPD}      & 3         \\
          \n tokens per type in \textit{pick different}           &\param{nTPD}       & 1         \\
          \n tokens in \textit{pick same}                         &\param{nTPS}       & 2         \\
          min \n available tokens in \textit{pick same}           &\param{minTPS}     & 4            
          \end{tabular}
          \caption{Parameters extracted from the game's rules.}
          \label{table:actions}
      \end{table}

    \subsection{1C2W and W2 Specifications}
      The two games introduced by the paper can be found in the folders "assets/1C2W2P/" and "assets/wacky2/" respectively. Each folder contains the following files:

      \begin{itemize}
        \item assets/default/decks/1.csv : list of cards in the cheap deck (cost, points, suit);
        \item assets/default/decks/2.csv : list of cards in the medium-cost deck (cost, points, suit);
        \item assets/default/decks/3.csv : list of cards in the expensive deck (cost, points, suit);
        \item assets/default/nobles/nobles.csv : list of nobles (cost and points);
        \item assets/default/parameters.json : game parameters for the engine.
      \end{itemize}

    \subsection{BMRH Hyperparameters}
      The BMRH hyperparameters are listed in Table \ref{tab:bmrh} while the values used can be found in Table \ref{tab:hpspace}.
        \begin{table}[h]
          \centering
          \begin{tabular}{l|c|l}
            \hline
            \textbf{Symbol}     &\textbf{Type}  & \textbf{Description}\\
            \hline
            $l$                 &integer        & sequence length \\
            $n$                 &integer        & sequences evaluated\\
            $usb$               &boolean        & if it uses shift buffer\\
            $mo$                &boolean        & if it has to mutate once\\
            $ms$                &integer        & mutation type\\
            $om$                &integer        & opponent model\\
            $ombs$              &double         & share of budget given to the opponent model\\
            $dcy$               &double         & probability of exponential decay\\
            $\mu$               &double         & mean of the gaussian mutation point\\
            $\sigma$            &double         & std dev of the gaussian mutation point
          \end{tabular}
          \caption{\label{tab:bmrh} Hyper-parameters of the BMRH agent.}
        \end{table}

        \begin{table}[h]
          \centering
          \begin{tabular}{l|l}
            \hline
            \textbf{Parameter}  &\textbf{Values}\\
            \hline
            $l$                 & $\{1,2,3,5,10,20\}$           \\        
            $n$                 & $\{20,50,100,200\}$           \\    
            $usb$               & $\{false,true\}$              \\   
            $mo$                & $\{false,true\}$              \\ 
            $ms$                & $\{0,1,2\}$                   \\
            $om$                & $\{0,1,2\}$                   \\
            $ombs$              & $\{0.005,0.01,0.02,0.05\}$    \\
            $dcy$               & $\{0.5,0.7,0.8,0.9\}$         \\
            $\mu$               & $\{0.0,0.1,0.3,0.5,0.75\}$    \\
            $\sigma$            & $\{0.5,1.0,2.0\}$                  
          \end{tabular}
          \caption{\label{tab:hpspace} BMRH hyperparameter space (total size of the space between parenthesis)}
        \end{table}

    \subsection{Event Type Id Mappings}
      The original event mapping can be found in Table \ref{table:events}.
      \begin{table}[h]
        \centering
        \resizebox{\columnwidth}{!}{
        \begin{tabular}{l|l|l|l|l}
          \textbf{State element} & \textbf{Event}           & \textbf{Who}  & $\mathbf{Type}^{id}$ & $\mathbf{Type}^{hc}$\\
          \hline
          Noble                  & place, take, receive     & $P_{i}$       & 7, 0, 14             & -1, -1, 3           \\
          Table's token          & increase, decrease       & $P_{i}$       & 1, 2                 & -1, -1              \\
          Table's joker          & increase, decrease       & $P_{i}$       & 3, 4                 & -1, -1              \\
          Table's card           & draw, place              & $P_{i}$       & 5, 6                 & -1, -1              \\
          Player's token         & increase, decrease       & $P_{i}$       & 8, 9                 & 0, -1               \\
          Player's joker         & increase, decrease       & $P_{i}$       & 10, 11               & 0, -1               \\
          Table's card           & reserve, hidden          & $P_{i}$       & 13, 12               & 2, 1                \\
          Player's points        & from card                & $P_{i}$       & 16                   & 4                   \\
          Player's points        & from noble               & E             & 17                   & 4                   \\
          Player's card          & buy                      & $P_{i}$       & 15                   & -1                  \\
        \end{tabular}}
        \caption{\label{table:events}List of all the events, $P_{i}$ is the $i$-th player, E for events triggered by the passive rule of \gameR's engine. When a state element has several events these are listed in the Event column separated by a comma and so are the relative ids.}
      \end{table}

  \section{Experiments}\label{sec:exp}
  	\subsection{Experiments Data}
      The data is saved in the folder \textit{data} and organised in subfolders: "data/[GAME]/vs [OPPONENT]/", where GAME is either SP2P, 1C2W or W2 and OPPONENT is only RND.
      However, given the submission limitations, we are uploading only the SP2P and W2 datasets since the other two wouldn't fit the 350MB limit.
  		The experiment produces a set of \textit{json} files containing useful information for the analysis of the results. Here's a description of the files:
  		
      \begin{itemize}
  			 \item \textit{agentSpace}: describes the hyperparameter space available of the BMRH algorithm;
			   \item \textit{behaviours}: lists the names of the behaviours;
			   \item \textit{bins}: for each behavioural dimension, lists the boundaries of the bins;
			   \item \textit{heuristicSpace}: brief description of the heuristic/mixing function that will be used by the agent;
			   \item \textit{opponents}: lists the opponent(s) against the agents will play;
			   \item \textit{p}: provides the information for running the experiment;
			   \item \textit{space}: lists all the points in the behavioural space at the end of the search.
			   \item \textit{spaceHistory}: lists all the points in the behavioural space found during the search together with the search iteration number of when they were found;
			   \item \textit{spaceSize}: lists the dimensions of the behavioural space;
			   \item \textit{summary}: short high-level list of game version, opponent, agent space and heuristic type;
			   \item \textit{support}: list of support metrics collected during the evaluation, these are not used in the behavioural space;
			   \item \textit{timing}: total running time and time per evaluation.
  		\end{itemize}

  	\subsection{Scripts}
  		The python scripts, that can be found in the homonym folder, are:

      \begin{itemize}
        \item \textit{bulk analysis}: runs the analysis on a selection of games and opponents (given the folder/file structure above);
        \item \textit{mapelites}: contains the code to load and analyse a single experiment;
        \item \textit{overlap}: contains the code to compare two experiments, it first checks that the behavioural spaces are compatible.
        \item \textit{space\_analysis}: contains the code to run a full single analysis and comparison between different search spaces on a single game.
      \end{itemize}

      Running the scripts produces the following \textit{csv} files: \textit{space}, \textit{spaceHistory}, \textit{spaceHistory\_binned}.
      The file \textit{space} has the fields:
      \begin{itemize}
        \item Behave\_D[$i$]: bin index for the $i$-th behaviour;
        \item fitness: average performance metric across several measurements;
        \item behaviours: list of values, one for each behavioural metric characterising the agent; 
        \item hyperparameters: hyperparameter configuration, algorithm and mixing function parameters are joined in this order.
      \end{itemize}

  		The file \textit{spaceHistory} has the fields:
      \begin{itemize}
        \item time: iteration number of when the point was found;
        \item behaviours: list of values, one for each behavioural metric characterising the agent;
        \item agentConfig: hyperparameters configuration of the algorithm;
        \item weights: weights configuration of the mixing function used by the algorithm;
        \item Behaviour[$i$]: list of $i$-th behaviour metric values, one for each evaluation;
        \item Support[$i$]: list of $i$-th support metric values, one for each evaluation;
      \end{itemize}
      
      The file \textit{spaceHistory\_binned} in addition to \textit{spaceHistory} has the field Behave\_D[i] which stores the coordinates in the behavioural space.
  	
    \subsection{Experiments Plots}
      The plots regarding a single experiment (given a game G, an opponent OPP, and an agent type AG) can be found at the path \mbox{"./out/[G]/[OPP] opponent/[AG]/"}.
      In addition to the plots described in the paper the script also produces a \textit{gif} file that shows the temporal evolution of all the 2D projections of the search space.

      Instead the plots for the comparison given two different search spaces AG1 and AG2 can be found at the path \mbox{"./out/[G]/vs[OPP]results/[AG1] vs [AG2]/"}

      In particular we want to highlight that the plots produced are straightforward to interpret and they further support the results highlighted in the paper. Unfortunately the amount of heatmaps produced makes it unfeasible to be analysed singularly.

      The typical structure is:
      \begin{itemize}
        \item \textit{[G]/[OPP] opponent/all\_space\_exploration.pdf}: plots that show what in the paper is called \textit{coverage};
        \item \textit{[G]/[OPP] opponent/all\_information\_quality.pdf}: plots that show  what in the paper is called \textit{convergence};
        \item \textit{[G]/[OPP] opponent/all\_performance\_distribution.pdf}: plots the distribution of performance in the range of [0,1] performance metric;
        \item \textit{[G]/[OPP] opponent/[AG]/plots/*}: all the 2D heatmaps of the final behavioural space;
        \item \textit{[G]/[OPP] opponent/[AG]/search\_progress.mp4}: temporal progress of the MAP-Elites search;
        \item \textit{[G]/[OPP] opponent/[AG1] vs [AG2]/[AG1] vs [AG2]/coverage/*}: all the 2D coverage comparison heatmaps of the final behavioural spaces of AG1 and AG2;
        \item \textit{[G]/[OPP] opponent/[AG1] vs [AG2]/[AG1] vs [AG2]/variation/*}: all the 2D performance delta comparison heatmaps of the final behavioural spaces of AG1 and AG2;
      \end{itemize}